\documentclass{article}

\usepackage{arxiv}

\usepackage[utf8]{inputenc} 
\usepackage[T1]{fontenc}    
\usepackage{hyperref}       
\usepackage{url}            
\usepackage{booktabs}       
\usepackage{amsfonts}       
\usepackage{nicefrac}       
\usepackage{microtype}      
\usepackage{lipsum}
\usepackage{graphicx}
\graphicspath{ {./images/} }

\usepackage{amsmath}
\usepackage{algorithm}
\usepackage{algpseudocode}

\algnewcommand\algorithmicforeach{\textbf{for each}}
\algdef{S}[FOR]{ForEach}[1]{\algorithmicforeach\ #1\ \algorithmicdo}

\title{Multimodal Story Generation on Plural Images}

\author{
 Jing Jiang \\
  The Cooper Union\\
  New York, NY 10003 \\
  \texttt{jiang6@cooper.edu} \\
}

\begin{document}
\maketitle
\begin{abstract}
Traditionally, the text generation models take in a sequence of text as input, and iteratively generate the next most probable word using pre-trained parameters. In this work, we propose the architecture to use images instead of text as the input of the text generation model, called StoryGen. In the architecture, we design a Relational Text Data Generator algorithm that relates different features from multiple images. The output samples from the model demonstrate the ability to generate meaningful paragraphs of text containing the extracted features from the input images.
\end{abstract}


\section{Introduction}
Machine learning based text generation models are excel in the task of generating coherent and human-understandable paragraphs of text. Gaining power from text generation models, a lot of applications improve themselves significantly, including machine translation, content completion, chat-bot implementation, etc. The core architecture behind the text generation models are Recurrent Neural Networks (RNNs). \par
RNNs are powerful in terms of learning and generating sequential data. It can be applied to applications in diverse domains including text, music and motion capturing data. In the task of text generation, it takes in and processes a sequence of text one step at a time, and predicts the next word. Traditionally, RNNs can not store the information from the input data for very long and behave in an unstable fashion. Gated Recurrent Units (GRU) and Long Short-term Memories (LSTM) are architectures that deal with this problem, by designing memory gates capable of storing and accessing data in a better way than traditional RNNs. Text generation tasks improves a lot in terms of coherency and human-readability using LSTMs. These LSTM based systems are categorized as supervised learning because they require a large corpus of text as dataset. Recently, many text generation models explore the design in an unsupervised way, and achieve state-of-the-art results. \par
Most of the case, text generation task relies on a given prompt in text form as input data. This work explores the possibility of using plural images as input data to generate a story based on the extracted features from the images. The process of extracting features from images and then generating semantically text is categorized as image captioning, a subset of multimodal learning. Multimodal learning aims to represent the joint representations of different modalities, such as text, images, and audio. Traditional image captioning models follow an encoder-decoder structure. The encoder is usually a convolutional neural network (CNN) which is capable of extracting features from the image. The decoder is a sequential model that decodes the extracted features from the encoder and restructure them into a sentence. Usally, the LSTM model is applied as the decoder sturcture in image captioning. By adding Attention mechanism to the decoder \cite{xu2015attend}, the models improve a lot in terms of interpretability, as the decoder can visualize its output when generating the description of the image. \par
The challenge to the task of generating text with multiple image as input is that the image captioning outputs from one image to another are not related. The existing image captioning models target for generating one sentence describing one image input. But for the task of story generation based on the extracted features on plural images, the extracted features from one of many image inputs must be stored for generating more than one sentences. Also, within one sentence of the story, there must features from more than one image inputs. To deal with this problem, we propose an algorithm that produces text that relates the features from plural image inputs one with another. The generated text can be used as training data for text generation models or directly as the sequential prompt for prediction of next words. \par
The contribution of this paper are the following:
\begin{itemize}
  \item We introduce an architecture, called StoryGen, that combines the power of image captioning models and text generation models, in order to generate paragraphs of text containing the extracted features of the input images.
  \item We design an algorithm, which can generate unsupervised text data that relates and links the extracted features from the images. The generated text can be used as training data or as prompt in text generation models.
\end{itemize} \par
Section 2 of this paper discusses the related works of this task. Section 3 describes the StoryGen architecture. Section 4 shows the proposed algorithm to generate text data. Section 5 shows the experiments and some samples of output using the StoyGen architecture.

\section{Related Works}
In this section we provide backgrounds for previous related works in the fields of text generation and image captioning.

\paragraph{Text Generation}
The approach to use of RNNs in the task of text generation came from \cite{graves2013generating}. This work proposed an architecture that is trained in a supervised fashion, where at each step, the model predict the most probable word. \par
A combination of pre-training and supervised fine-tuning is commonly used in supervised learning based language task. The learning of word vectors used as inputs to task-specfic models is introduced by \cite{mikolov2013distributed}. \par
The methods above require supervised training in performing of a task. There are works that design approaches using unsupervised learning, such as \cite{schwartz-etal-2017-story} and \cite{radford2017learning}. A variation of text generation model via Generative Adversarial Networks are presented by \cite{guo2017long}.

\paragraph{Image Captioning}
The first neural networks based approach for multimodal image captioning task was proposed by \cite{kiros2014unifying}. This work introduced a multimodal log-bilinear architecture biased by features from the input image. The use of Recurrent Neural Network in the decoder part is proposed by \cite{mao2014deep}. This model was then improved by replacing a traditional RNN with an LSTM model in works from \cite{vinyals2014tell} and \cite{donahue2014longterm}. \par

The works above represent images as a single feature vector from the top layer of a pre-trained convolutional network. The notion of joint embedding space for ranking and generation is brought up by \cite{karpathy2014deep}. This work introduced an architecture that has a R-CNN object detection algorithm that learns to score similarity between sentence and the image using the outputs from a bidirectional RNN. This model first learn the detector based on a multi-instance learning framework, following by a language model trained on captions, and finally feeding into a rescoring mechanism from a joint embedding space of image and text. \par

The concept of Attention was introduced in the task of image captioning by \cite{xu2015attend}. This work proposed an architecture that superimpose the Attention layers to the RNN decoder layer. By doing so, the model learned to attend to abstract concepts that goes beyond objectness in previous works. Also, by bringing in Attention mechanism, this architecture improves the interpretability of image captioning models a lot.

\section{Architecture}
In this section, we describe the general pipeline of story generation architecture using plural images as inputs. This section is divided into three subsections. The first subsection introduces the overall pipeline of StoryGen architecture. The second subsection introduces the detailed image captioning procedure within the first stage of story generation. The relational text generator is described in detail in the third subsection. The last subsection describes the text generation procedure based on the output from the relational text data generator. 

\begin{figure} 
    \makebox[\textwidth][c]{\includegraphics[width=0.9\textwidth]{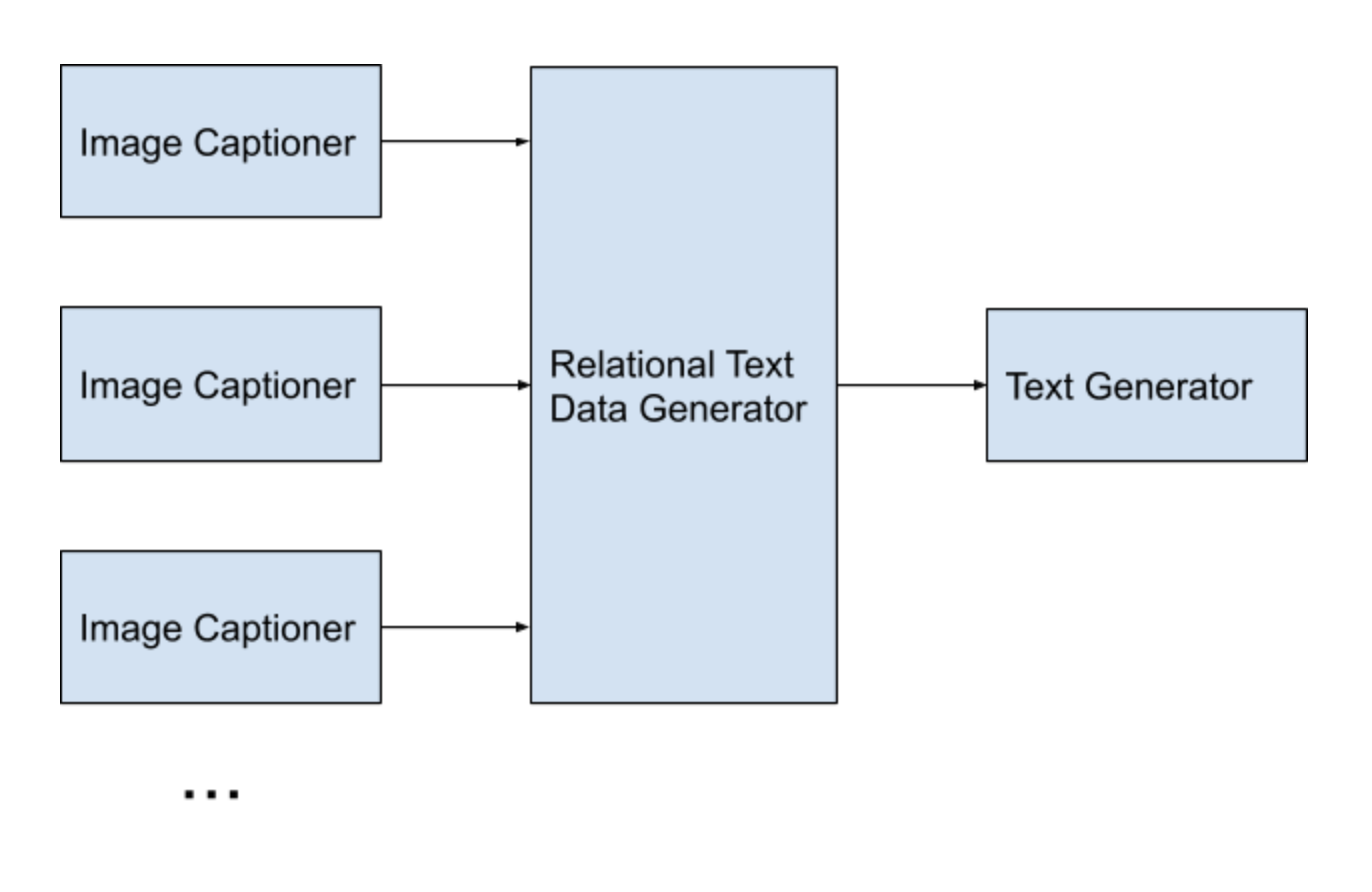}}%
    \caption{StoryGen overall pipeline}
\end{figure}

\subsection{Overall Pipeline}
Our model take multiple images as input and generate a paragraph of story with features extracted from the images. The model can be divided into three stages. The first stage consists of a image captioner that separately captions one input image. The second stage is the relational text data generator that deal with the lack of relationship between the extracted features. The generator takes the individual captions from Stage 1 as input and outputs unsupervised sentences that consist of extracted features as subjects, objects, and adjectives. The last stage is the text generator, which generate the story paragraph based on the output sentences generated by the text data generator. Figure 1 describes the overall procedure of StoryGen architecture.

\begin{figure} 
    \makebox[\textwidth][c]{\includegraphics[width=0.9\textwidth]{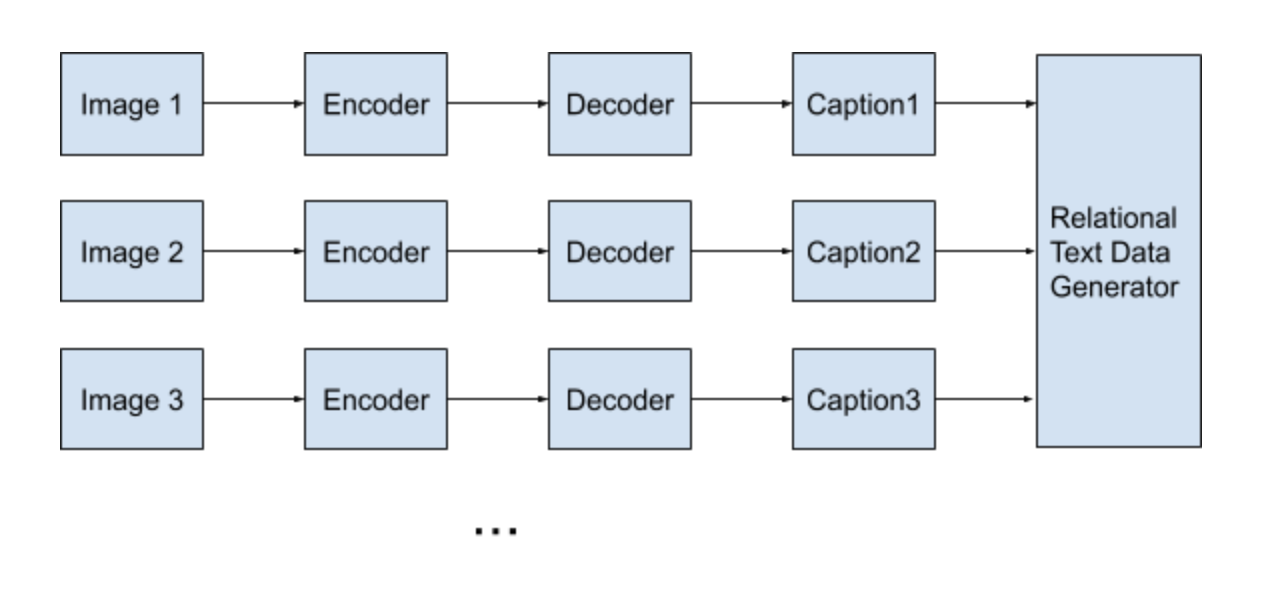}}%
    \caption{Detailed view of The Image Captioner pipeline}
\end{figure}

\subsection{Image Captioner}
The image captioning mechanism used in StoryGen architecture is the Show, Attend and Tell model introduced by \cite{xu2015attend}. The Image Captioner follows an Encoder-Decoder architecture. Figure 2 provides a detailed pipeline of the Image Captioner. The Image Captioner is trained on Flickr8K dataset, which has around 8000 jpg images and 40000 captions to annotate the corresponding images. During training, we use the train-test split provided by \cite{karpathy2014deep}. \par

The encoder is an convolutional neural network that extract the features of the single input image. The encoder used in the Image Captioner is a pre-trained 101 layered Residual Network (ResNet-101) model. This model can capture the essence of an input image very well. The fine-tuning of the Image Captioner is enabled, so that when it is training, the Encoder network can improve itself using the output of the Decoder network. \par

During encoding of a single input image. The Encoder progressively create smaller and smaller feature representations of the original image. And as each subsequent feature representation goes more into depth of the Encoder network, there is a greater number of channels. The final encoding tensor produced by our ResNet-101 Encoder has a size of 14x14 with 2048 channels.\par

When decoding the encoded tensor, the input feeds itself into a Recurrent Neural Network and generate sequential data. We use an LSTM model with soft Attention mechanism. During decoding, the network attends to different part of the image at different point in the sequence, and generate the most probable caption using Beam Search.

\subsection{Relational Text Data Generator}

\begin{algorithm}
\caption{Relational Text Data Generator}\label{alg:mlc}
\begin{algorithmic}[1]
\State \textbf{Input}: captions of images
\State \textbf{Output}: corpus of text that relates the features of images
\State \textbf{Hyperparameter 1}: the rate that the subject, object, or adjective is replaced $\alpha \in [0, 1]$
\State \textbf{Hyperparameter 2}: number of iterations $k$

\State \textbf{Initialization}: Extract the subject, objects, and adjectives from the captions to form a list of potential word replacements: $L$
\For{$iteration \gets 1$ to $k$}
\State Randomly choose a verb (see, smell, hear …) to form an SVO sentence
\State Generate some sentences using the Text Generation model with the formed sentence as customized prompts
\State Randomly replace some subjects, objects, and adjectives in $L$, with the rate of $\alpha$
\State Append the replaced sentence to the output

\EndFor
\end{algorithmic}
\end{algorithm}
The Relational Text Data Generator runs an algorithm that takes in the captions of the images and outputs a text corpus with the extracted features from the images.\par
As shown in Algorithm 1, the Relational Text Data Generator initialize a parameter $\alpha$ to denote the rate of replacement, and an iteration number $k$. Before the iteration, the Relational Text Data Generator extracts the features from the images, categorizes them as subjects, objects, and adjectives, and store them into list of potential replacements.\par
During each iteration, the Relational Text Data Generator first use the a subject from the list and a caption input to form a SVO sentence. Then use this sentence as a prompt to generate one more sentence using a pre-trained text generator. Finally, the Relational Text Data Generator randomly replace some of the subjects, objects and adjectives from the generated sentence and append this new sentence to the output text corpus.\par

\subsection{Text Generator}
After the generation of unsupervised text using the Relational Text Data Generator, the data feeds into the Text Generator either as training data or as customized prompts. \par
If the data is to be used as training data, the number of iteration should be set as a large number, since many text generation models requires large amount of text corpus as training data. If the data is to be used as a customized prompt at inference stage, large number of iteration will not be required. \par
The model to provide the results is GPT-2 from OpenAI, a model with state-of-the-art output in the field of text generation. GPT-2 is a large transformer-based language model with 1.5 billion parameters, trained on a dataset of 8 million web pages. GPT-2 has the objective of predicting the next word, given all of the previous sentences as the prompt. The diversity of the dataset used by OpenAI causes the goal to contain naturally occurring demonstrations of many tasks across many language domains. GPT-2 is a direct scale-up of the previous GPT, with more than 10X the parameters and trained on more than 10X the amount of data.\par

Since OpenAI does not provide the code for training the GPT-2 model, the output from the Relational Text Data Generator is used to provide customized prompts.

\section{Results}
\begin{figure} 
    \makebox[\textwidth][c]{\includegraphics[width=0.9\textwidth]{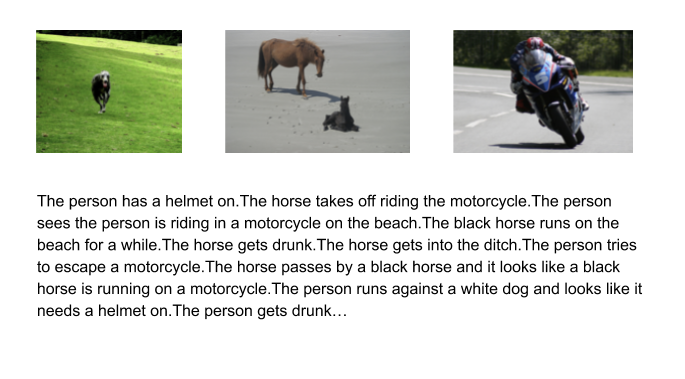}}%
    \caption{Sample output from StoryGen}
\end{figure}
We have conducted the experiments with a $\alpha$ of 0.5 and an iteration number of 3. The output of the Relational Text Data Generator is used as the prompts of the Text Generator model. The model successfully generate meaningful paragraphs. \par
As Figure 4 shows, the features from the images (dog, horse, person, helmet, motorcycle) are correctly identified and put together into a story in the generated texts.

\section{Conclusion}
In this work, we proposed StoryGen, which explores the possibility to use images instead of texts in text generation tasks. We identify the difficulty of the lack of relationship between input images and design the algorithm to generate unsupervised text corpus that relates the extracted features from the images. The sample output from StoryGen shows that the features of the images are related in a coherent way. \par
Future works of this direction include designing a more compact architecture, because the current architecture follows a three-step pipeline that rely on different models in different task domains. Also, works that improve semantics in the generated paragraphs are worth exploring.

\bibliographystyle{unsrt}  
\bibliography{references} 

\end{document}